\documentclass[10pt,twocolumn,letterpaper]{article}

\usepackage{cvpr}      

\definecolor{cvprblue}{rgb}{0.21,0.49,0.74}
\usepackage[pagebackref,breaklinks,colorlinks,allcolors=cvprblue]{hyperref}

\def\paperID{15} 
\def\confName{CVPR}
\def\confYear{2026}

\title{
Bridging Foundation Models and ASTM Metallurgical Standards for Automated Grain Size Estimation from Microscopy Images\\
}

\author{
Abdul Mueez \qquad Shruti Vyas \\ [0.5ex]
University of Central Florida \\
{\tt\small \{abdul.mueez, shruti\}@ucf.edu}
}

\begin{document}
\maketitle

\begin{abstract}
Extracting standardized metallurgical metrics from microscopy images remains challenging due to complex grain morphology and the data demands of supervised segmentation. To bridge foundational computer vision with practical metallurgical evaluation, we propose an automated pipeline for dense instance segmentation and grain size estimation that adapts Cellpose-SAM to microstructures and integrates its topology-aware gradient tracking with an ASTM E112 Jeffries planimetric module. We systematically benchmark this pipeline against a classical convolutional network (U-Net), an adaptive-prompting vision foundation model (MatSAM) and a contemporary vision-language model (Qwen2.5-VL-7B). Our evaluations reveal that while the out-of-the-box vision-language model struggles with the localized spatial reasoning required for dense microscopic counting and MatSAM suffers from over-segmentation despite its domain-specific prompt generation, our adapted pipeline successfully maintains topological separation. Furthermore, experiments across progressively reduced training splits demonstrate exceptional few-shot scalability; utilizing only two training samples, the proposed system predicts the ASTM grain size number ($G$) with a mean absolute percentage error (MAPE) as low as 1.50\%, while robustness testing across varying target grain counts empirically validates the ASTM 50-grain sampling minimum. These results highlight the efficacy of application-level foundation model integration for highly accurate, automated materials characterization. Our project repository is available at \url{https://github.com/mueez-overflow/ASTM-Grain-Size-Estimator}.

\end{abstract}

\section{Introduction}
\label{sec:intro}

Grain size, morphology and distribution play a crucial role in determining mechanical properties. Generally, metals with fine, evenly distributed grains exhibit greater strength and ductility compared to those with coarse, irregular grains \cite{schempp2013influence, do2007effect, wang1995effect}. Beyond strength and ductility, grain characteristics also influence other properties, including electrical and thermal conductivity, corrosion resistance and wear resistance \cite{heo2006influence, uddin2010effect, ali2022computational}.

Considering the singular importance of grain structure, its rapid measurement and quantification remain a key objective of materials analysis \cite{yan2017grain, yakout2018review}. Traditionally, the determination of grain boundaries from cross-sectional images relied on manual inspection, which is highly accurate but extremely time-consuming, prone to operator fatigue and vulnerable to human error and inconsistency \cite{dengiz2005grain}. To alleviate this burden, computational techniques relying on classic image processing algorithms—such as manual thresholding, watershed algorithms and gradient-based edge detection—have been widely employed \cite{catania2022new, dengiz2005grain}. However, these traditional techniques often suffer from well-known inefficiencies, including severe sensitivity to image quality, heavy reliance on manual parameter tuning and a lack of standardization across varying samples \cite{frazier2014metal, tan2020microstructure}.

These segmentation challenges become even more pronounced with the advent of modern manufacturing techniques, such as metal additive manufacturing (AM) \cite{adam20183d, chen2009modeling, herriott2019multi}. AM processes, which depend on layer-by-layer deposition and anisotropic sintering behavior, often lead to non-traditional and complex grain structures \cite{adam20183d, chen2009modeling, herriott2019multi}. In powder-based 3D-printed and sintered materials, full densification is rarely achieved, leaving residual pores that can interfere with grain boundary recognition \cite{olevsky2018field}. Furthermore, non-optimized etching often erodes grain boundaries or connects adjacent pores, producing highly unclear and discontinuous boundaries that standard computational algorithms fail to accurately segment \cite{satterlee2025robust}.

To address the limitations of traditional image processing, data-driven machine learning (ML) and deep learning models have increasingly been adopted for microstructural analysis. While ML models can automate the process and handle complex features like impurities and polishing scratches, they are critically dependent on large, expertly annotated training datasets for their success \cite{perera2021optimized}. Acquiring high-quality metallographic training data is often tedious, expensive and severely restricts the rapid deployment of these models for novel material systems \cite{perera2021optimized}.

To overcome this data-dependency bottleneck, researchers are increasingly looking toward foundation models. Foundation models such as the Segment Anything Model (SAM) possess distinctive characteristics that make them highly adaptable \cite{kirillov2023segment}. Because they are pretrained on extremely large datasets, they learn representations with strong inductive biases \cite{pachitariu2025cellpose}. These biases can significantly improve out-of-distribution generalization, particularly when fine-tuning with limited domain-specific data \cite{hendrycks2020pretrained, fort2021exploring}. Despite these advantages, the base SAM architecture has design choices that make it poorly suited for the dense image segmentation required for metallographic analysis \cite{pachitariu2025cellpose}. Therefore, this study introduces a fine-tuned Cellpose-SAM approach specifically adapted for metallographic analysis. By overcoming the dense segmentation limitations of the base architecture, this approach successfully delineates complex, microscopic grain boundaries to directly predict the ASTM grain size number ($G$) in strict accordance with the ASTM E112-25 planimetric procedure \cite{ASTM_E112_25}. The ASTM grain size number is a standardized, dimensionless metric used to quantify grain density \cite{ASTM_E112_25}; importantly, it scales logarithmically, meaning a higher $G$ value corresponds to a finer grain structure, which typically indicates enhanced mechanical strength and ductility \cite{callister2020materials}. By automating the extraction of this standardized value, our approach provides a direct, highly accurate bridge between advanced foundation models and practical metallurgical evaluation.

In summary, the main contributions of this work are as follows:


\begin{itemize}
    \item We systematically benchmark Cellpose-SAM and MatSAM against a U-Net baseline and evaluate the zero-shot counting capability of Qwen2.5-VL, establishing the limitations of text-based visual decoders on dense microstructural domains.
    \item We demonstrate that domain-adapted Cellpose-SAM significantly outperforms all baselines, maintaining topological separation without the over-segmentation of MatSAM and exhibiting strong few-shot scalability from as little as 2 training samples.
    \item We implement a fully automated Jeffries planimetric pipeline that predicts the ASTM grain size number ($G$) with an MAPE as low as 1.50\% and empirically validate the ASTM 50-grain sampling minimum across varying target grain counts.
\end{itemize}

\section{Related Work}
\label{subsec:related_work}

The application of foundation models to domain-specific segmentation tasks has seen rapid adoption. Recently, the ImageGrains 2.0 framework demonstrated the viability of fine-tuning Cellpose-SAM for geoscientific applications, successfully segmenting macroscopic sediment particles such as fluvial gravel and sand from optical and X-ray computed tomography (XR-CT) imagery \cite{mair2026imagegrains}. 

While ImageGrains 2.0 established that Cellpose-SAM's biological inductive biases can transfer to macroscopic clastic grains, the fine-tuning of this architecture for microscopic metallographic grain boundaries remains largely unexplored. Traditional automated metallographic studies have predominantly relied on traditional thresholding and watershed algorithms \cite{dengiz2005grain}, which often struggle to generalize across varying etching qualities, surface scratches, or annealing twins \cite{satterlee2025robust, warren2024grain}. 

To overcome these limitations, recent approaches have increasingly adopted deep convolutional neural networks (CNNs), particularly U-Net variants, to improve segmentation robustness under challenging microstructural conditions \cite{patrick2023automated,satterlee2025robust,warren2024grain}. For instance, Patrick et al. \cite{patrick2023automated} demonstrated the efficacy of a U-Net model paired with specialized post-processing algorithms to successfully segment grain boundaries in bright-field transmission electron microscopy (TEM) images, effectively navigating the complex diffraction contrasts inherent to the modality. A critical bottleneck in these supervised deep learning approaches, however, remains the scarcity of high-quality, expertly annotated metallographic training data. To address this, Warren et al. \cite{warren2024grain} developed a hybrid dataset of manually segmented AM 316L micrographs augmented with synthetic Voronoi-tessellated grain structures.


To bypass this data-dependency bottleneck entirely, researchers have also explored zero-shot foundation models for materials science. Most notably, MatSAM \cite{matsam} introduced a prominent unsupervised and training-free framework for material microstructures. MatSAM adaptively generates prompt points for different microscopy images, fuses the centroid points of coarsely extracted regions of interest (ROI) and native grid points and integrates corresponding post-processing operations for quantitative characterization. While innovative in its prompt engineering, relying purely on zero-shot inference without domain-specific weight updates can lead to topological inconsistencies in flawed or densely packed microstructures.


\section{Methodology}

\begin{figure*}[t]
    \centering
    \includegraphics[width=\textwidth]{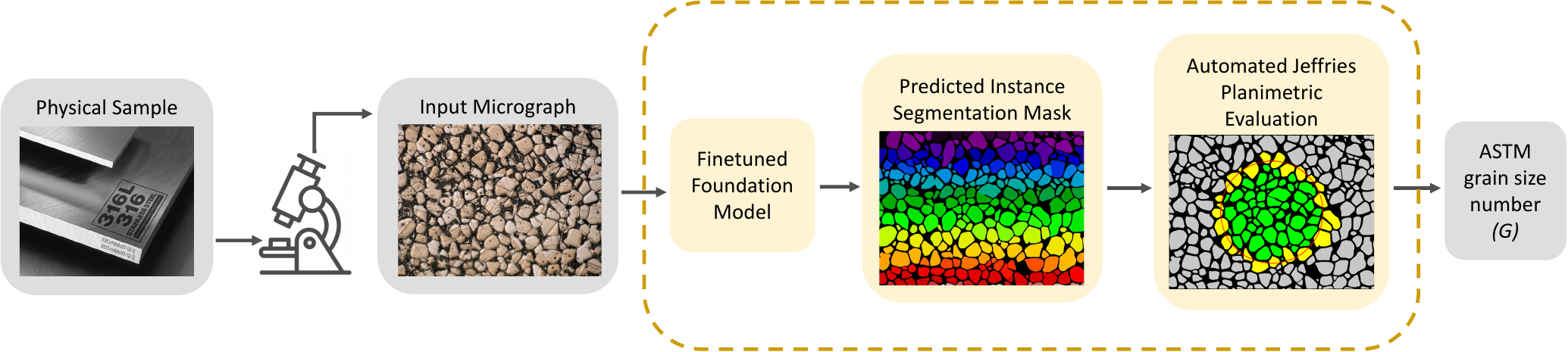}
    \caption{Proposed automated metallographic analysis workflow. A microscopy image with known magnification is processed by a finetuned Cellpose-SAM model pipeline to output a dense segmentation mask. Subsequent image processing utilizing the Jeffries' method directly calculates the ASTM grain size number (G).}
    \label{fig:workflow_pipeline}
\end{figure*}

\subsection{Dataset Preparation and Image Reconstruction}
\label{subsec:dataset_preparation}

This study utilizes a dataset of ExOne Stainless Steel 316L microscopy images \cite{warren2023voronoi_grains_kaggle, warren2024grain}. This specific material system was selected because additively manufactured (AM) alloys present highly challenging, porous microstructures that rigorously test segmentation limits. Crucially, this dataset was chosen because it uniquely satisfies two strict prerequisites for our automated pipeline: it provides high-quality, expert-annotated ground truth masks and it can be cross-referenced with a physical spatial scale. The primary dataset version utilized for training and evaluation contains 480 localized, high-magnification image patches with manual annotations. However, during its creation, the region containing the physical scale bar was removed. To recover this vital information, we cross-referenced an alternative version of the dataset that retained the original scale bar region \cite{warren2023voronoi_grains_kaggle_with_scale}. Because the standard ASTM E112-25 Jeffries method mandates a known physical area to calculate grain density \cite{ASTM_E112_25}, this spatial calibration is an absolute requirement for end-to-end automation.

\begin{figure}[t]
    \centering
    \begin{subfigure}{0.48\columnwidth}
        \centering
        \includegraphics[width=0.23\linewidth]{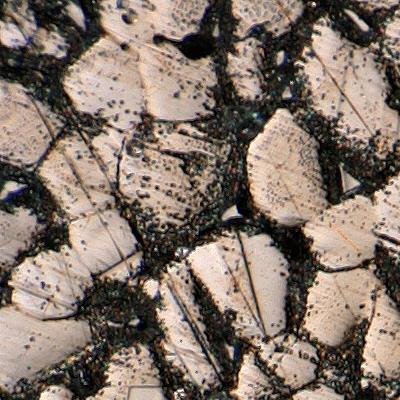}\hfill
        \includegraphics[width=0.23\linewidth]{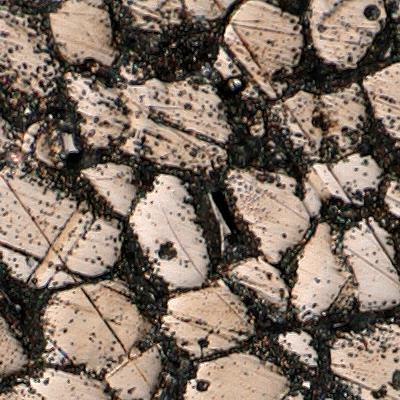}\hfill
        \includegraphics[width=0.23\linewidth]{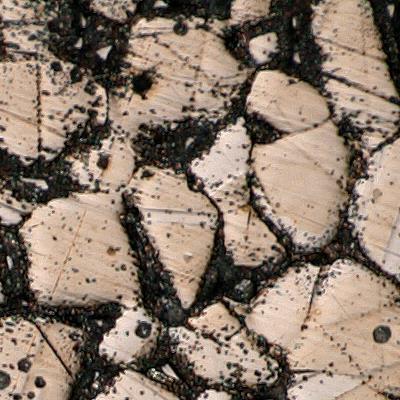}\hfill
        \includegraphics[width=0.23\linewidth]{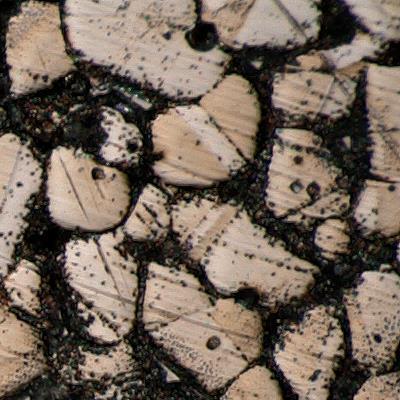}\vspace{0.02\linewidth}
        
        \includegraphics[width=0.23\linewidth]{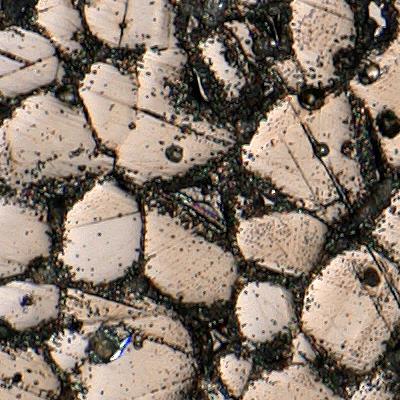}\hfill
        \includegraphics[width=0.23\linewidth]{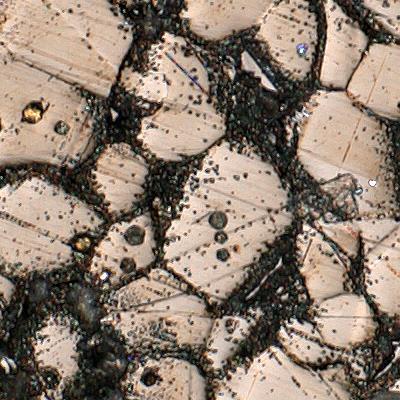}\hfill
        \includegraphics[width=0.23\linewidth]{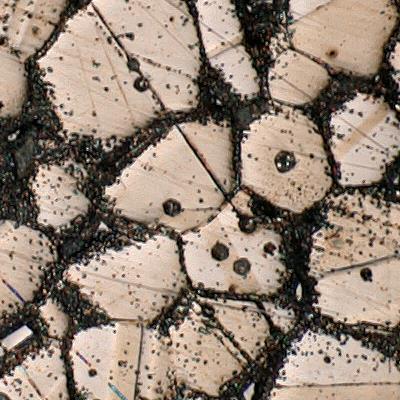}\hfill
        \includegraphics[width=0.23\linewidth]{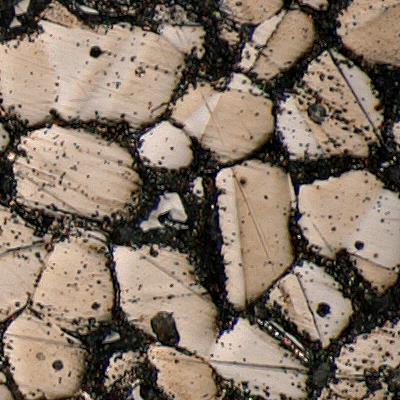}\vspace{0.02\linewidth}
        
        \includegraphics[width=0.23\linewidth]{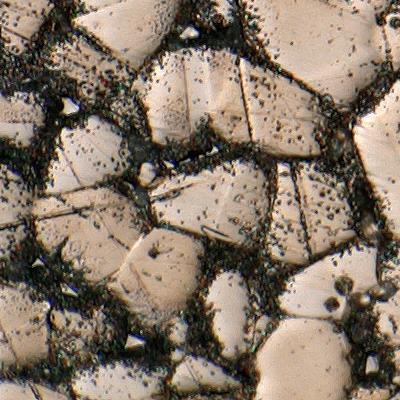}\hfill
        \includegraphics[width=0.23\linewidth]{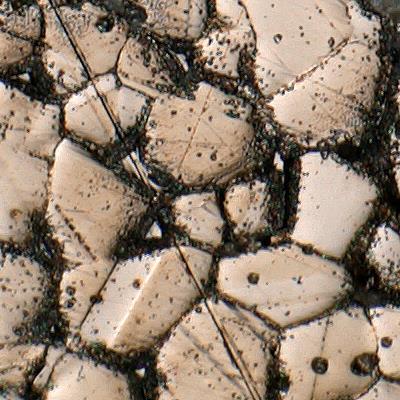}\hfill
        \includegraphics[width=0.23\linewidth]{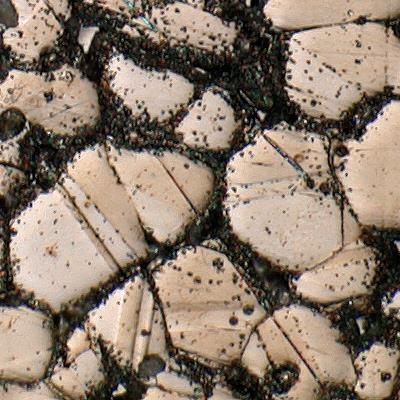}\hfill
        \includegraphics[width=0.23\linewidth]{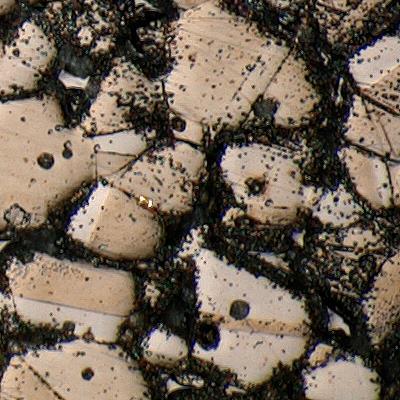}
        \caption{Original patches}
        \label{fig:patches_grid}
    \end{subfigure}
    \hfill 
    \begin{subfigure}{0.48\columnwidth}
        \centering
        \includegraphics[width=\linewidth]{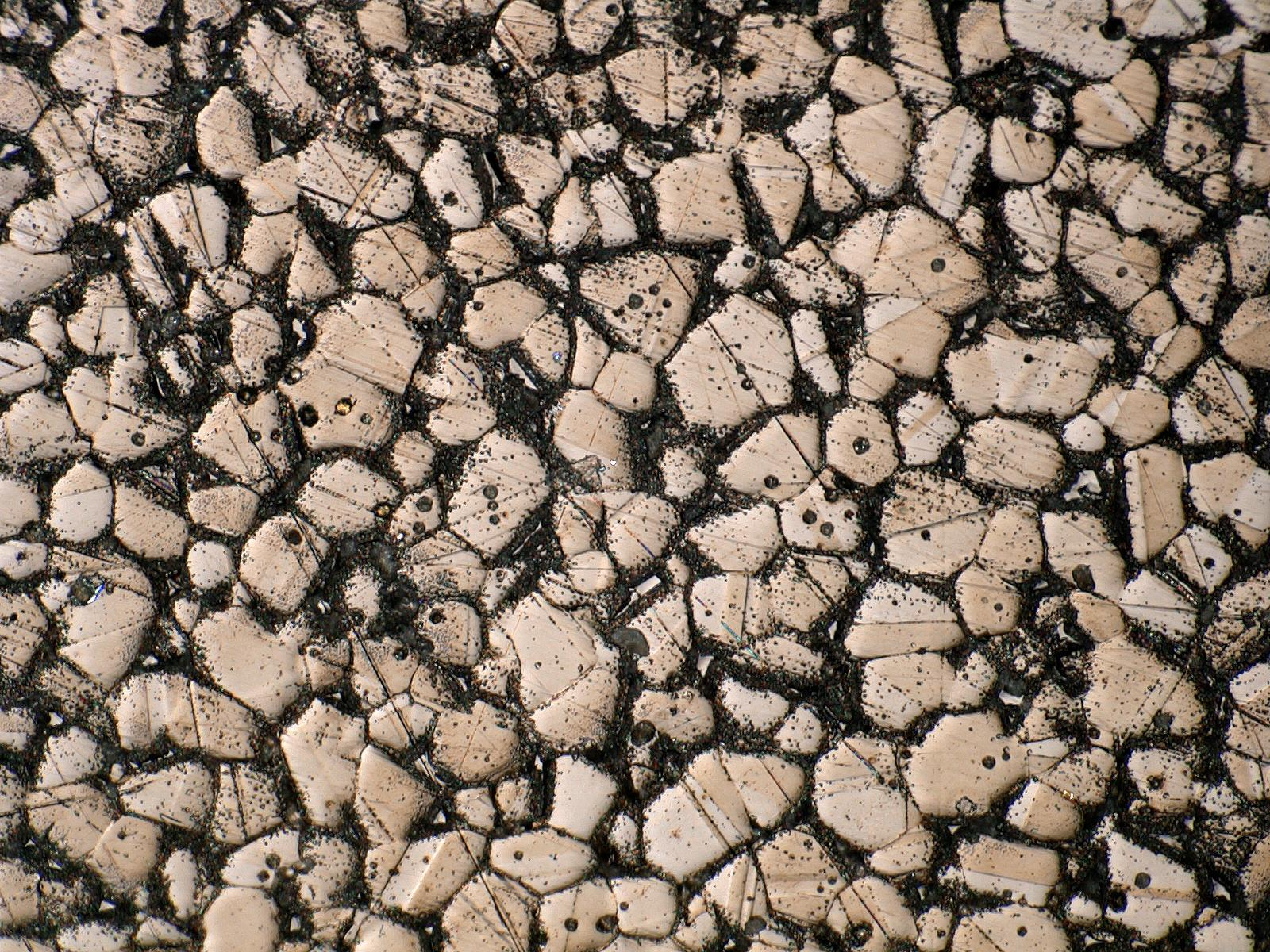}
        \caption{Reconstructed}
        \label{fig:stitched_result}
    \end{subfigure}
    
    \caption{Dataset reconstruction process. (a) The original dataset provides localized microscopy patches that possess a field of view too small to encompass the 50 grains required by ASTM E112-25. (b) The programmatic stitching of 12 contiguous patches into a single, comprehensive micrograph, enabling the standard Jeffries planimetric procedure.}
    \label{fig:reconstruction_process}
\end{figure}

To meet these standardization requirements, we developed an automated pipeline that parses spatial coordinate metadata from filenames to stitch groups of 12 adjacent patches, aggregating the 480 patches into 40 full-field micrographs and corresponding label masks (Fig.~\ref{fig:reconstruction_process}). The expanded field of view permits inscription of a test circle large enough to satisfy ASTM E112-25 requirements.

\subsection{The Jeffries Planimetric Procedure}
\label{subsec:jeffries_method}

The Jeffries planimetric method serves as the standard referee procedure for determining the average grain size of materials \cite{ASTM_E112_25}. The method evaluates the density of grains contained within a test area of known size. 

\begin{figure}[t]
    \centering
    \includegraphics[width=0.5\columnwidth]{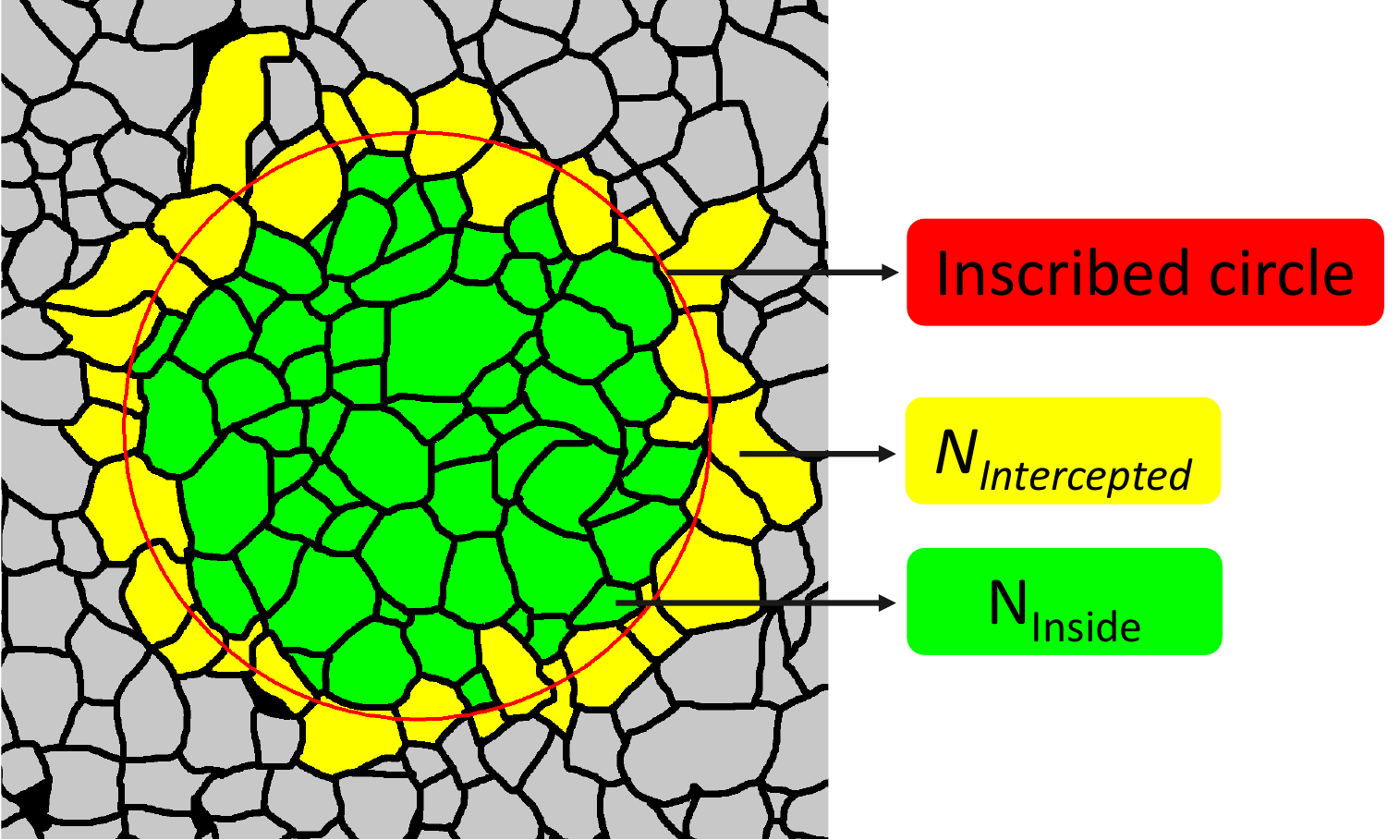}
    \caption{Application of the Jeffries planimetric method. Whole grains lie entirely within the test circle (green); intersecting grains are counted as one-half (yellow).}
    \label{fig:jeffries_circle}
\end{figure}

An inscribed circle, typically representing a $5000\,\text{mm}^2$ area, is overlaid on the micrograph at a magnification that ensures at least 50 grains are captured \cite{ASTM_E112_25}. As shown in Fig.~\ref{fig:jeffries_circle}, grains situated entirely within the circle are counted as whole units, while those bisected by the test boundary are counted as half units. The number of grains per mm$^{2}$ at $1\times$ magnification, $N_A$, is calculated as:

\begin{equation}
N_A = f\left(N_{\text{Inside}} + \frac{N_{\text{Intercepted}}}{2}\right)
\end{equation}

where $f$ is the Jeffries multiplier. For a standard $5000\,\text{mm}^2$ area, the multiplier is defined by the image magnification $M$:

\begin{equation}
f = 0.0002\,M^2
\end{equation}

This $N_A$ value (grain density at $1\times$ magnification) directly dictates the dimensionless ASTM grain size number $G$:

\begin{equation}
G = 3.321928\,\log_{10}(N_A) - 2.954
\end{equation}

\subsection{Mechanisms of the Cellpose-SAM Framework}
\label{subsec:architecture}

Cellpose-SAM transitions away from traditional U-Net semantic segmentation \cite{satterlee2025robust, ronneberger2015u, patrick2023automated, warren2024grain} and leverages the zero-shot feature extraction capabilities of the Segment Anything Model (SAM) \cite{kirillov2023segment} and topological gradient-tracking mechanism of Cellpose \cite{pachitariu2025cellpose}. 

Standard SAM relies on user prompts to guide its mask decoder. Cellpose-SAM bypasses this limitation by discarding the SAM decoder entirely. It retains only the pretrained SAM Vision Transformer (ViT) image encoder, which provides highly generalized, robust inductive biases. 

The embeddings generated by the SAM encoder are fed into a transposed convolution layer to predict three continuous dense fields: vertical spatial flows, horizontal spatial flows and overall cell probabilities. Cellpose's gradient-tracking algorithm then parallelizes across these vector flow fields, tracking pixels back to their respective dynamic centers to establish non-overlapping instance masks. This architectural fusion achieves dense, prompt-free segmentation while capitalizing on SAM's superior representational depth.

\subsection{Mask Processing and Instance Labeling}
\label{subsec:mask_processing}

Because the Cellpose gradient-tracking mechanism requires distinct instance labels rather than binary edge maps, the original ground truth masks required morphological transformation.

First, to isolate the grain interiors from the boundary lines, the input masks ($I_{mask}$) were binarized at a threshold of $\tau = 128$. The isolated grain interiors, $S_{int}$, are defined by extracting the regions bounded by these edges:

\begin{equation}
S_{int}(x,y) = \begin{cases} 1 & \text{if } I_{mask}(x,y) < \tau \\ 0 & \text{otherwise} \end{cases}
\end{equation}

Crucially, to prevent gradient-tracking failure caused by adjacent instances ``bleeding'' into one another, morphological erosion was applied to separate any subtly touching grains. Using a disk-shaped structuring element $D$ of radius 1, the binarized interiors were updated as:

\begin{equation}
S_{int}' = S_{int} \ominus D
\end{equation}

where $\ominus$ represents the erosion operator. Following this separation, a size-exclusion filter eliminated any residual noise or etching artifacts by removing connected components with an area $A(C) < 200$ pixels. Finally, a connected component labeling (CCL) algorithm assigned a unique integer identifier to each distinct interior. This ensures the model clearly resolves individual cell centers with a single-pixel background buffer. The final masks were saved as 16-bit TIFFs; because the dense metallic microstructure frequently approaches or exceeds the 255-instance limit of standard 8-bit encoding, this expanded bit-depth is required to safely assign a unique integer to every grain without risking integer overflow.

\subsection{Implementation Details and Fine-Tuning Strategy}
\label{subsec:training}

To adapt the base foundation model for microstructural analysis, we utilized an aggressive data augmentation pipeline to combat the inherent scarcity of annotated metallographic data. Because the smallest training split contained only 2 images, each original micrograph was augmented 30 times. This extensive augmentation was necessary to provide the model with sufficient gradient diversity per epoch and prevent rapid memorization.


The pipeline combined geometric distortions (random flips, rotations, elastic and grid distortions), photometric adjustments (CLAHE, random gamma shifts and Gaussian noise) and coarse dropout to simulate morphological variation, etching artifacts and force learning of broader topological continuity.

Training was conducted for 100 epochs using the Cellpose optimization framework. Instead of exhaustive hyperparameter tuning—often computationally inefficient and prone to over-fitting on extremely small datasets—hyperparameters were selected using standard stabilization heuristics for small datasets. A low learning rate of $1 \times 10^{-6}$ was used to prevent divergence of the training loss, while a batch size of 2 enabled frequent updates. A weight decay of $0.2$ provided strong $L_{2}$ regularization to mitigate overfitting. All experiments were implemented in PyTorch and run on a single NVIDIA RTX A5500 GPU.

\subsection{Automated Calculation of $N_A$ via the Jeffries Method}
\label{subsec:automated_jeffries}


An automated pipeline programmatically executes the Jeffries method by dynamically inscribing a circle enclosing a target of 60 internal grains---a 10-grain buffer above the ASTM minimum \cite{ASTM_E112_25}---with the effect of varying this target analyzed in Section~\ref{subsubsec:autonomous_robustness}. For baseline evaluations, this exact dynamically sized circle is superimposed onto the corresponding prediction mask. This strict spatial alignment isolates the model's counting error from natural microstructural spatial variance, guaranteeing a direct, head-to-head evaluation of the $N_{\text{Inside}}$ and $N_{\text{Intercepted}}$ metrics.


Within this fixed test boundary, grains are classified purely via Euclidean distance from the image center: grains whose maximum radial pixel distance falls within the circle's radius are counted as $N_{\text{Inside}}$, while those whose minimum radial distance is within but maximum exceeds it are counted as $N_{\text{Intercepted}}$.

To compute the true number of grains per unit area ($N_A$) in industrial compliance with ASTM E112-25, a precise spatial calibration was established. As detailed in Section~\ref{subsec:dataset_preparation}, a reference scale bar was extracted from the dataset version containing the scale information \cite{warren2023voronoi_grains_kaggle_with_scale}. Analysis of this scale bar determined that $100\,\mu\text{m}$ is equivalent to 226 pixels (a calibration factor of $2.26\,\text{px}/\mu\text{m}$).





Rather than applying this spatial calibration to the entire rectangular field of view---which would violate the geometric principles of the circular Jeffries method---the algorithm calculates the exact physical area of the dynamically inscribed test circle for each individual micrograph. For an optimal test circle with a pixel radius of $r$, the true physical area $A_{\text{circle}}$ (in $\text{mm}^2$) is calculated as $A_{\text{circle}} = \pi r^2 / (c^2 \times 10^6)$, where $c = 2.26\,\text{px}/\mu\text{m}$ is the calibration factor and $10^6$ converts the resulting area from square micrometers to square millimeters.

Under ASTM E112-25, the Jeffries multiplier $f$ is mathematically defined as $M^2/A$, where $M$ is the magnification and $A$ is the test figure area in $\text{mm}^2$ \cite{ASTM_E112_25}. Because the automated pipeline computes the true physical area directly from the microstructural pixel scale, the magnification factor is effectively $M = 1$. The multiplier therefore simplifies to the inverse of the circle's physical area, creating a dynamic multiplier ($f_{\text{dynamic}} = 1 / A_{\text{circle}}$) that adapts to the specific test area drawn for each image.

By applying this dynamically calculated multiplier, the algorithm seamlessly normalizes the localized grain count to a standard $1\times$ magnification density. Within the evaluation test set, the dynamic test circles required to capture the target 60 internal grains ranged in physical area from $0.090$ to $0.117\,\text{mm}^2$. Consequently, the automated pipeline applied corresponding $f_{\text{dynamic}}$ multipliers ranging from 8.55 to 11.14. This adaptive normalization ensures that regardless of the test circle's fluctuating pixel radius, the final $N_A$ metric remains mathematically rigorous, allowing for a highly accurate, standardized calculation of the ASTM grain size number ($G$) across the entire dataset.

\section{Results and Analysis}
\label{subsec:results}

Model performance was evaluated across progressive training data splits—5\% (2 images), 10\% (4 images), 25\% (10 images), 50\% (20 images) and 75\% (30 images)—to test few-shot scalability. Splits were constructed cumulatively, with each smaller dataset a strict subset of the next larger. The remaining 10 full-field micrographs (25\% of the dataset) were reserved as a fixed evaluation test set across all experiments. The proposed Cellpose-SAM pipeline was benchmarked against two distinct approaches: a standard 4-stage U-Net semantic segmentation architecture—trained from scratch on the 75\% data split using a combined BCE-Dice loss and morphological post-processing—and the unsupervised MatSAM framework, which utilized a ViT-Large encoder, adaptive grid prompting and strict confidence thresholds for zero-shot inference. Results were assessed using standard instance segmentation metrics to measure topological integrity, alongside strict, domain-specific metallurgical calculations.

\subsection{Standard Instance Segmentation and Boundary Quality}

Average Precision (AP) and Boundary F1 scores indicate the topological alignment and pixel-level precision of the models. The strict mAP (0.50--0.95) metric evaluates the integrity of individual grain instances across increasingly demanding Intersection over Union (IoU) thresholds.

\begin{table}[htbp]
\centering
\caption{Instance Segmentation and Boundary Quality Metrics. The proposed Cellpose-SAM framework across progressive training splits is compared against U-Net and MatSAM baselines. \textbf{Best scores are bolded.}}
\label{tab:segmentation_metrics}
\resizebox{\columnwidth}{!}{%
\begin{tabular}{@{}lcccc@{}}
\toprule
\textbf{Model / Split} & 
\shortstack{{AP} \\ {@ 0.50}} & 
\shortstack{{mAP} \\ {(0.50--0.95)}} & 
\shortstack{{Boundary} \\ {F1}} & 
\shortstack{{Count} \\ {Error}} \\ 
\midrule
U-Net (75\%)        & 0.4717 & 0.2796 & \textbf{0.8305} & -98.3 \\
MatSAM (Zero-Shot)  & 0.2688 & 0.1224 & 0.6559 & +151.6 \\ 
\midrule
Cellpose-SAM (Zero-Shot) & 0.4006 & 0.2089 & 0.5022 & -125.0 \\
Cellpose-SAM (5\%)  & 0.6457 & 0.3691 & 0.7704 & \textbf{-17.6} \\
Cellpose-SAM (10\%) & 0.6664 & 0.3884 & 0.7905 & -24.0 \\
Cellpose-SAM (25\%) & 0.6820 & 0.4058 & 0.8131 & -32.3 \\
Cellpose-SAM (50\%) & \textbf{0.6856} & 0.4107 & 0.8161 & -35.7 \\
Cellpose-SAM (75\%) & 0.6836 & \textbf{0.4150} & 0.8236 & -30.2 \\ 
\bottomrule
\end{tabular}%
}
\end{table}

\begin{figure*}[htbp]
    \centering
    \includegraphics[width=\textwidth]{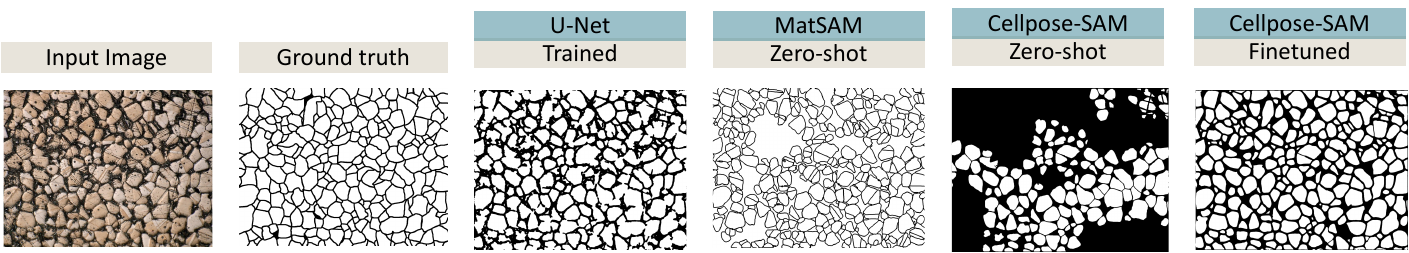}

    \caption{Qualitative comparison of instance segmentation across architectures. Zero-shot Cellpose-SAM under-segments significantly, but fine-tuning on just 2 samples (5\% split) yields superior topological alignment. U-Net merges adjacent instances despite training on 75\% of data. MatSAM (ViT-Large) severely over-segments porous regions despite its full adaptive prompting pipeline.}
    \label{fig:model_comparison}
\end{figure*}

As demonstrated in Table~\ref{tab:segmentation_metrics}, utilizing the Cellpose-SAM architecture in a purely zero-shot capacity yields inadequate results for dense metallographic analysis. The base model struggles to identify continuous edges, resulting in a poor Boundary F1 of 0.5022 and massive under-segmentation, missing an average of 125 grains across the entire reconstructed field of view. However, exposing the model to merely 5\% of the training data triggers a massive performance leap. The AP@0.50 jumps by over 24 percentage points to 0.6457 and the count error shrinks to just -17.6. This rapid convergence confirms that SAM’s inherited inductive biases provide an exceptionally strong foundation, requiring only minimal domain-specific fine-tuning to successfully adapt to microscopic boundary detection. As training data increases, Boundary F1 steadily improves (0.8236 at 75\%) while count error slightly worsens---a sign of evolving conservatism. The model becomes adept at tracing high-confidence edges but skips faint, ambiguous boundaries caused by poor etching or porosity, merging distinct grains and registering topological errors.

This critical difference between pixel-level and instance-level evaluation is starkly illustrated by the baseline comparisons. U-Net achieves a slightly higher Boundary F1 (0.8305) but its mAP (0.2796) and count error ($-98.3$) are considerably worse: while it accurately classifies edge pixels, it fails to close topologically distinct contours, bleeding adjacent grains together. Conversely, MatSAM's adaptive prompting overreacts to the texture, porosity and scratches of 3D-printed metals, causing severe over-segmentation (+151.6 count error) and the lowest mAP (0.1224).

\subsection{ASTM Jeffries Planimetric Evaluation}

To validate the true utility of the segmentations, the resulting masks were translated into practical metallurgical parameters via the automated Jeffries pipeline. Because the test set remains constant across all primary evaluations using the baseline 60-grain target, the ground truth (GT) averages are fixed: 60.3 for internal grains ($N_{\text{Inside}}$), 31.9 for intercepted grains ($N_{\text{Intercepted}}$), a grain density ($N_A$) of 752.7 and an ASTM grain size ($G$) of 6.60.

\begin{table*}[htbp]
\centering
\caption{Automated Jeffries Planimetric Evaluation Metrics. The baseline Ground Truth (GT) averages are listed in the column headers. The proposed Cellpose-SAM model's average predictions (Pred) and Mean Absolute Percentage Errors (MAPE) across progressive training data splits are compared against U-Net (trained on a 75\% split) and MatSAM (ViT-Large) baselines. The notations represent the number of internal grains ($N_{\text{Inside}}$), the number of intercepted grains ($N_{\text{Intercepted}}$), the equivalent whole grains per unit area or grain density ($N_A$) and the dimensionless ASTM Grain Size number ($G$). \textbf{Lowest MAPE in each column is bolded.}}
\label{tab:jeffries_metrics}
\resizebox{0.8\textwidth}{!}{%
\begin{tabular}{@{}lcccccccc@{}}

\toprule
& \multicolumn{2}{c}{$N_{\text{Inside}}$ (GT: 60.3)} & \multicolumn{2}{c}{$N_{\text{Intercepted}}$ (GT: 31.9)} & \multicolumn{2}{c}{$N_A$ (GT: 752.7)} & \multicolumn{2}{c}{$G$ (GT: 6.60)} \\ 
\cmidrule(lr){2-3} \cmidrule(lr){4-5} \cmidrule(lr){6-7} \cmidrule(l){8-9} 
Model / Split & Pred & MAPE (\%) & Pred & MAPE (\%) & Pred & MAPE (\%) & Pred & MAPE (\%) \\ 
\midrule

U-Net (75\% Train) & 43.3 & 28.19 & 26.2 & 18.96 & 554.5 & 26.05 & 6.15 & 6.78 \\
MatSAM (Zero-shot) & 104.8 & 73.76 & 40.2 & 29.10 & 1226.7 & 63.81 & 7.30 & 10.61 \\

\midrule

Cellpose-SAM (Zero-Shot)  & 38.4 & 36.31 & 17.7 & 45.01 & 125.7 & 38.11 & 3.92 & 16.70 \\
Cellpose-SAM (5\% Train)  & 57.2 & \textbf{7.80}  & 27.0 & 15.42 & 695.6 & \textbf{8.20}  & 6.48 & \textbf{1.88} \\
Cellpose-SAM (10\% Train) & 56.4 & 8.79  & 26.7 & 16.47 & 685.4 & 9.72  & 6.46 & 2.24 \\
Cellpose-SAM (25\% Train) & 55.1 & 8.95  & 27.6 & 13.39 & 676.4 & 9.78  & 6.45 & 2.28 \\
Cellpose-SAM (50\% Train) & 55.1 & 9.28  & 27.4 & 13.90 & 676.1 & 9.79  & 6.44 & 2.30 \\
Cellpose-SAM (75\% Train) & 56.6 & 10.45 & 27.9 & \textbf{12.41} & 692.7 & 10.12 & 6.48 & 2.34 \\

\bottomrule
\end{tabular}%
}
\end{table*}

Table~\ref{tab:jeffries_metrics} contextualizes the models' accuracy using the Mean Absolute Percentage Error (MAPE). To ensure a rigorous comparison, the MAPE evaluates the average magnitude of the predictive error relative to the expert-annotated ground truth (GT). As reflected by these error rates, the U-Net model's tendency to merge instances results in a severely under-counted $N_{\text{Inside}}$ average of 43.3, driving its $N_A$ error to 26.05\% and its $G$ error to 6.78\%. MatSAM's fragmentation of the microstructures inflates its $N_A$ error to a massive 63.81\%, falsely indicating a much finer grain structure and resulting in a $G$ prediction error of 10.61\%.

The proposed Cellpose-SAM pipeline, however, provides estimations that sit comfortably within industrial tolerances. Even at the minimal 5\% training split, it predicts an average $N_{\text{Inside}}$ of 57.2, tightly aligning with the 60.3 ground truth. Crucially, while the MAPE for the raw grain density ($N_A$) hovers between 8\% and 10\% for the fine-tuned models, the MAPE for the final ASTM Grain Size ($G$) is exceptionally low. The pipeline achieves the lowest MAPE of 1.88\% at the 5\% training split and remains under 2.35\% across all other fine-tuned iterations. This stability is a direct result of the logarithmic scaling of the ASTM grain size formula. Moderate counting variances in the linear $N_A$ domain are heavily compressed, meaning that despite the model's slight topological under-segmentation on ambiguous boundaries, the integrated pipeline yields highly reliable, standardized metallurgical classifications that outperform current CV approaches.


Notably, the best metallurgical accuracy ($N_A$ MAPE 8.20\%, $G$ MAPE 1.88\%) occurs at the minimal 5\% split. This reflects an evolving conservatism: with minimal data, the model leverages SAM's inductive biases to close boundaries on faint evidence, whereas additional training on ambiguous artifacts causes it to restrict predictions to high-confidence edges only, merging grains and slowly inflating MAPE. The sole exception is $N_{\text{Intercepted}}$ MAPE, which monotonically improves with training (best 12.41\% at 75\%), as additional data corrects the 5\% model's tendency to spuriously seal perimeter-crossing grains into interior predictions.

\subsubsection{Robustness of the Autonomous Pipeline}
\label{subsubsec:autonomous_robustness}

A key architectural consideration in the Jeffries pipeline is how the inscribed test circle is determined. In the primary evaluation (Table~\ref{tab:jeffries_metrics}), the test circle is dynamically sized on the ground truth (GT) mask to enclose a fixed target of 60 internal grains and subsequently superimposed onto the prediction mask at the same position and radius---a design choice that ensures a direct, head-to-head spatial comparison. While this constitutes the methodologically rigorous approach for benchmarking count accuracy against a known GT geometry, it raises a practical question: does the pipeline's performance depend on access to the ground truth mask for circle placement?

To answer this, we conducted two complementary experiments on the Cellpose-SAM model fine-tuned on the 5\% split. First, we varied the target grain count from 10 to 100 in increments of 10, measuring how the MAPE evolves as the enclosed region grows---using the GT-superimposed circle in each case. Second, we compared this against an autonomous setup in which the GT and prediction masks each independently draw their own circles by applying the target-count algorithm directly to their respective segmentations, with no shared geometric information between them. Table~\ref{tab:combined_scale_metrics} presents the results of both experiments.

\begin{table*}[htbp]
\centering
\caption{GT-Free Pipeline Robustness vs.\ GT-Derived Baseline (Cellpose-SAM, 5\% split). The GT-Derived configuration superimposes the circle drawn on the GT mask onto the prediction mask; the GT-Free configuration independently draws a circle on each mask using the same target count, requiring no ground truth access at inference. Both are evaluated across varying target grain counts. GT values reflect the true local grain density within each sampling region. \textbf{Lowest MAPE in each column is bolded.}}
\label{tab:combined_scale_metrics}
\resizebox{0.87\textwidth}{!}{%
\begin{tabular}{@{} lcc | ccccc | ccccc @{}}
\toprule
\multicolumn{3}{c|}{{Ground Truth (GT)}} & \multicolumn{5}{c|}{{Superimposed Circle (GT-Derived)}} & \multicolumn{5}{c}{{Independent Circle (GT-Free)}} \\
\cmidrule(r){1-3} \cmidrule(lr){4-8} \cmidrule(l){9-13}
Count & $N_A$ & $G$ & Count & $N_A$ & MAPE (\%) & $G$ & MAPE (\%) & Count & $N_A$ & MAPE (\%) & $G$ & MAPE (\%) \\
\midrule
10  & 811.7 & 6.70 & 9.4  & 694.8 & 17.18 & 6.47 & 4.12 & 10  & 712.0 & 17.26 & 6.50 & 4.07 \\
20  & 773.9 & 6.63 & 18.6 & 690.7 & 13.20 & 6.47 & 3.13 & 20  & 710.7 & 14.19 & 6.51 & 3.29 \\
30  & 761.0 & 6.61 & 28.8 & 688.0 & 10.51 & 6.47 & 2.46 & 30  & 709.2 & 11.60 & 6.51 & 2.64 \\
40  & 757.1 & 6.61 & 37.7 & 686.9 &  9.51 & 6.47 & 2.21 & 40  & 683.9 &  9.74 & 6.46 & 2.27 \\
50  & 741.7 & 6.58 & 48.5 & 682.8 &  7.91 & 6.46 & 1.81 & 50  & 692.8 &  7.54 & 6.48 & 1.71 \\
60  & 752.7 & 6.60 & 57.2 & 695.6 &  8.20 & 6.48 & 1.88 & 60  & 694.2 &  7.58 & 6.48 & 1.76 \\
70  & 747.1 & 6.59 & 67.7 & 693.0 &  7.98 & 6.48 & 1.81 & 70  & 692.5 &  7.71 & 6.48 & 1.78 \\
80  & 751.9 & 6.60 & 76.1 & 691.9 &  8.07 & 6.48 & 1.88 & 80  & 691.0 &  7.92 & 6.48 & 1.83 \\
90  & 748.9 & 6.59 & 85.6 & 695.7 & \textbf{7.22} & 6.48 & \textbf{1.69} & 90  & 701.4 & \textbf{6.50} & 6.50 & \textbf{1.50} \\
100 & 756.7 & 6.61 & 95.6 & 700.1 &  7.67 & 6.49 & 1.79 & 100 & 699.9 &  8.27 & 6.50 & 1.88 \\
\bottomrule
\end{tabular}%
}
\end{table*}

\paragraph{Effect of Target Grain Count.}
As the target grain count increases from 10 to 100, both the GT reference and model error undergo clear transitions. At low counts, the GT $N_A$ is unstable---reaching 811.7\,grains/mm$^2$ at a target of 10 grains before plateauing to $N_A \approx 741$--$757$ beyond 50 grains---a sampling artifact amplified by the inversely proportional $f_{\text{dynamic}}$ multiplier. The model's predicted $N_A$, by contrast, remains narrow around 683--712\,grains/mm$^2$ across all counts, a consequence of conservative under-segmentation producing a spatially homogeneous prediction field. At low targets, this counting shortfall compounds the unstable GT reference, driving $G$ MAPE to $\sim$4\%. As the target grows, both effects attenuate: the GT reference stabilizes and the logarithmic scaling of $G$ absorbs residual under-counting, causing $G$ MAPE to fall steeply to $\sim$1.7--1.8\% by 50 grains and plateau in the 1.50\%--1.88\% band from 50 to 100. This validates the ASTM E112-25 minimum of 50 grains, below which spatial sampling bias and stochastic counting variance dominate the error budget.

\paragraph{GT-Independence of the Autonomous Pipeline.}
The most practically significant finding is the near-identical performance of the GT-Free and GT-Derived configurations. Across all ten target counts, the absolute difference in $G$ MAPE is at most 0.19 percentage points, with the GT-Free approach marginally superior at several counts---most notably at the 60-grain target (1.76\% vs.\ 1.88\%). This directly refutes any concern that pipeline accuracy depends on GT circle geometry. The pipeline is therefore fully self-contained: at deployment, a practitioner supplies only the raw micrograph and a target grain count ($\geq$50 to satisfy ASTM requirements) and the system independently inscribes a test circle and computes $G$ without any ground truth dependency.

\subsection{Zero-Shot Generalization on Diverse Microstructures}
\label{subsec:zero_shot_generalization}

\begin{figure}[htbp]
    \centering
    \includegraphics[width=\columnwidth]{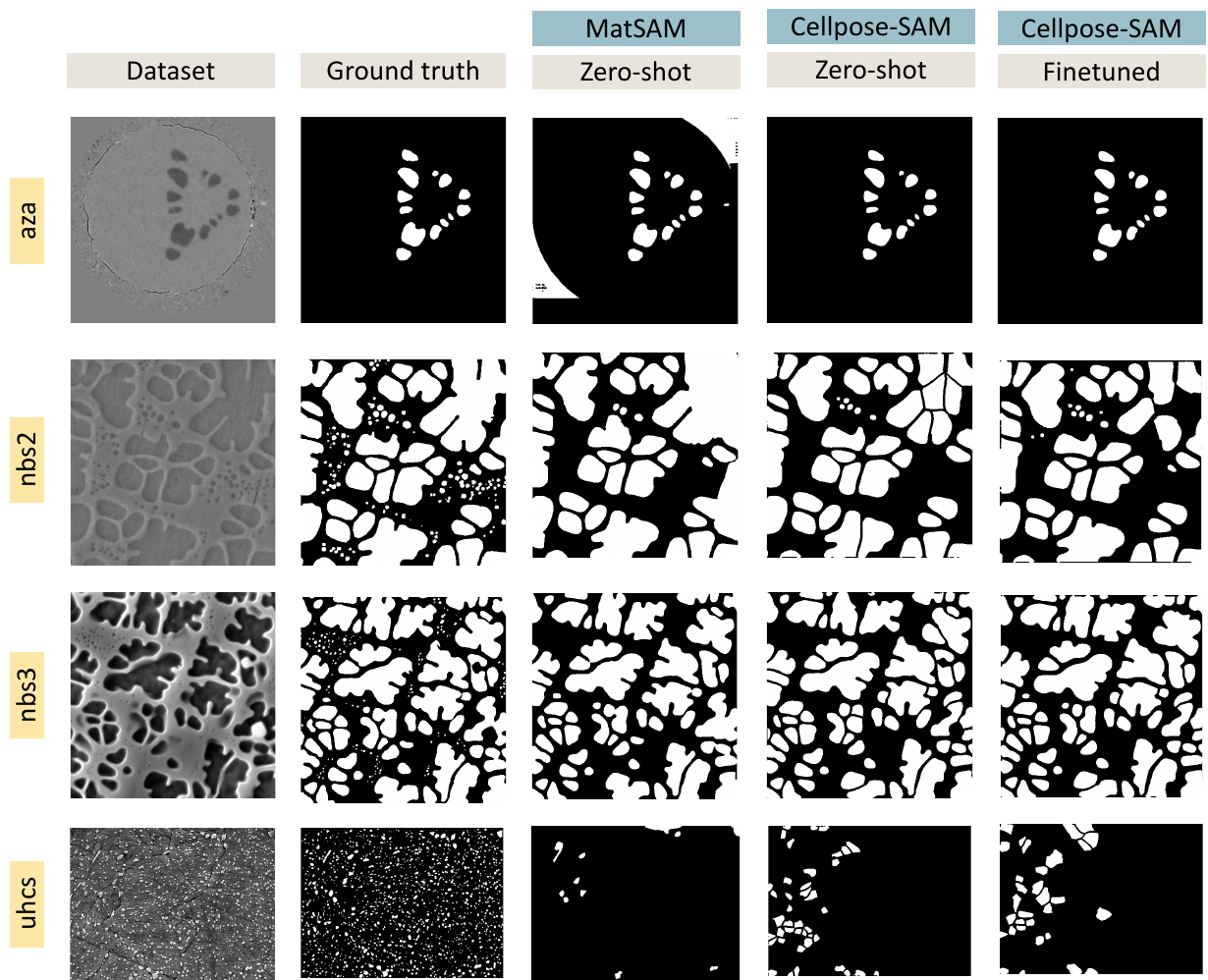}
    \caption{Qualitative zero-shot performance comparison on four diverse alloy datasets (AZA, NBS-2, NBS-3 and UHCS) \cite{matsam_dataset}. The figure displays the first test image from each dataset alongside its ground truth mask. Predictions from the zero-shot Cellpose-SAM architecture and the training-free MatSAM framework highlight differences in out-of-distribution generalization.}
    \label{fig:QA}
\end{figure}

Grain structures exhibit profound morphological diversity across different material classes. To further evaluate the out-of-distribution robustness and zero-shot generalization capabilities of the architectures, we conducted a qualitative assessment across four diverse, external microstructural datasets \cite{matsam_dataset} (Fig.~\ref{fig:QA}): AZA (a polycrystalline AlZn alloy with dark central precipitates), NBS-2 and NBS-3 (single-crystal nickel-based superalloys featuring highly variable tertiary precipitates) and UHCS (an ultra-high-carbon steel featuring fine, spherical cementite precipitates distributed along a rough alloy crack).

Visual analysis reveals differences in zero-shot adaptability. On the AZA dataset, MatSAM generates a massive, erroneous circular mask bounding the entire region, whereas Cellpose-SAM accurately isolates the target central precipitates. On NBS-2/3, both models identify the primary structures, but Cellpose-SAM demonstrates tighter topological adherence to the complex, branching dendritic shapes. Finally, the UHCS dataset proves highly challenging for unsupervised methods; MatSAM under-segments the image, failing to detect the vast majority of the fine spherical cementite precipitates. Notably, fine-tuning on the 5\% stainless steel split does not degrade out-of-distribution robustness; the fine-tuned model closely matches the zero-shot baseline while capturing additional grains across diverse morphologies.


\subsection{Alternative Approach with Vision-Language Models (VLMs)}
\label{subsec:vlm_approach}

\begin{figure}[htbp]
    \centering
    \includegraphics[width=0.6\columnwidth]{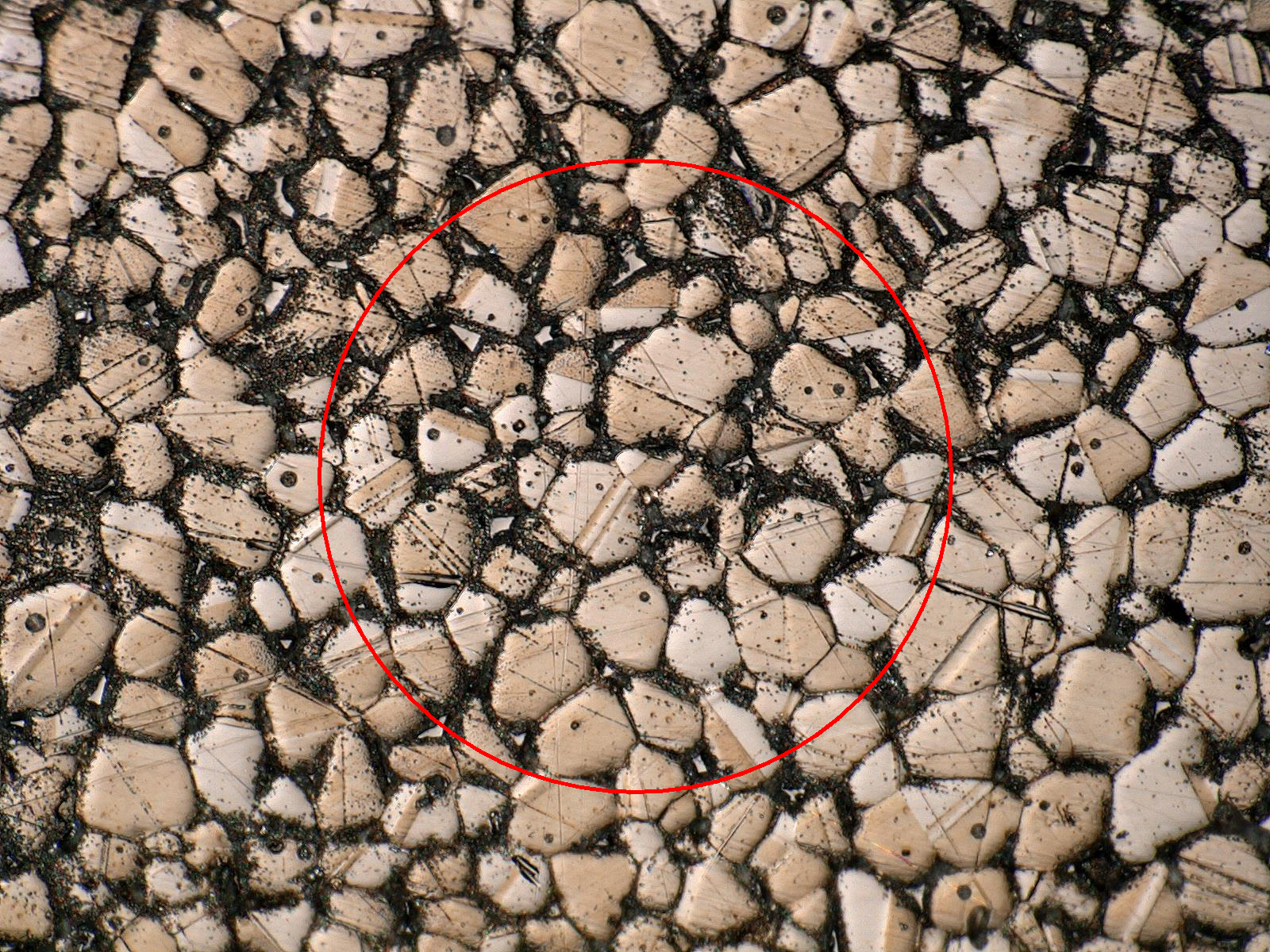} 
    \caption{Example of the visual input to the Vision-Language Model: the raw micrograph overlaid with a red test circle matching the ground truth mask.}
    \label{fig:vlm_input}
\end{figure}

We also explored a purely zero-shot alternative using a Vision-Language Model (VLM). Specifically, we deployed Qwen2.5-VL-7B-Instruct to directly execute the Jeffries planimetric grain count without any intermediate segmentation steps. As shown in Fig.~\ref{fig:vlm_input}, each test image was presented as the raw micrograph overlaid with an inscribed red circle matching the GT mask from the 60-grain evaluation in Table~\ref{tab:jeffries_metrics}, with the model tasked to directly output $N_{\text{Inside}}$ and $N_{\text{Intercepted}}$.

Despite strong general visual reasoning, the VLM failed entirely, hallucinating identical outputs for every image ($N_{\text{Inside}} = 0$, $N_{\text{Intercepted}} = 5$). This failure highlights a current limitation of text-based visual decoders like the tested Qwen2.5-VL-7B model when applied out-of-the-box to this domain. While they excel at global image comprehension and macroscopic object detection, this specific zero-shot setup lacked the high-resolution, localized spatial attention required for dense, microscopic counting tasks. This observation aligns with broader findings in recent literature, which demonstrate that contemporary VLMs frequently struggle with dense, instance-level counting \cite{qharabagh2024lvlm, li2024naturalbench, vo2025vision}.

\section{Conclusion}
\label{sec:conclusion}

This study introduces a vision foundation model-based approach for automated segmentation of complex microscopic metallic grain boundaries. By adapting foundational representations alongside topological gradient-tracking, the proposed method successfully overcomes the dense segmentation limitations and over-segmentation issues inherent to standard unsupervised foundation models. Specifically, we demonstrate that the fine-tuned Cellpose-SAM architecture achieves exceptional few-shot scalability; even when fine-tuned on merely 5\% of the available data, the pipeline reliably predicts the standardized ASTM grain size number ($G$) with a minimal Mean Absolute Percentage Error (MAPE) as low as 1.50\%. Crucially, our robustness analysis across varying target grain counts empirically validates the ASTM 50-grain sampling minimum and confirms that the pipeline functions entirely autonomously, eliminating any reliance on ground-truth geometric priors during deployment. Through this fully autonomous approach, we move beyond traditional pixel-level evaluations to successfully bridge foundational computer vision capabilities directly with standardized industrial evaluation. This approach proves especially necessary given the limitations of alternative architectures, including the catastrophic failure of a zero-shot VLM for direct planimetric counting.

{
    \small
    \bibliographystyle{ieeenat_fullname}
    \bibliography{main}
}


\end{document}